\documentclass{article}
\usepackage{ijcai18}

\usepackage{times}
\usepackage{helvet}
\usepackage{courier}
\usepackage{graphicx}
\usepackage{color,xcolor}
\usepackage{pgf,tikz}

\usetikzlibrary{arrows,automata}
\usepackage{verbatim}

\usepackage{amssymb,amsmath}
\usepackage[small]{caption}
\usepackage[linesnumbered,ruled,vlined]{algorithm2e}

\newtheorem{THEOREM}{Theorem}
\newenvironment{theorem}{\begin{THEOREM} }%
                        {\end{THEOREM}}
\newtheorem{LEMMA}[THEOREM]{Lemma}
                      {\end{LEMMA}}
\newtheorem{PROPOSITION}[THEOREM]{Proposition}
\newenvironment{prop}{\begin{PROPOSITION} }
                      {\end{PROPOSITION}}
\newtheorem{COROLLARY}[THEOREM]{Corollary}
                          {\end{COROLLARY}}
\newtheorem{EXAMPLE}{Example}
\newenvironment{example}{\begin{EXAMPLE} \rm}%
                            {\end{EXAMPLE}}

\newtheorem{DEFINITION}{Definition}
\newenvironment{definition}{\begin{DEFINITION} \rm }
                            {\end{DEFINITION}}

\newtheorem{PROBLEM}{Problem}

\newcommand{\ie}{\textit{i.e.}}
\newcommand{\eg}{\textit{e.g.}}

\newcommand{\add}{\texttt{add}}
\newcommand{\swap}{\texttt{swap}}
\newcommand{\drop}{\texttt{drop}}

\newcommand{\Rz}{R_{\geq 0}}
\newcommand{\confChange}{\mathit{confChange}}
\newcommand{\free}{\mathit{free}}
\newcommand{\unlocker}{\mathit{unlocker}}
\newcommand{\NULL}{NULL}
\newcommand{\wv}{w_{\mathcal{V}}}
\newcommand{\we}{w_{\mathcal{E}}}

\newcommand{\lastStepImproved}{\mathit{lastStepImproved}}
\newcommand{\hash}{\mathit{hash}}

\title{Advancing Tabu and Restart in Local Search for Maximum Weight Cliques}

\author{Yi Fan$^{1,2,3}$\thanks{Corresponding author}, Nan Li$^4$, Chengqian Li$^5$, Zongjie Ma$^3$, Longin Jan Latecki$^4$, Kaile Su$^{2,3}$\\
$^1$Guangxi Key Laboratory of Trusted Software, Guilin University of Electronic Technology, Guilin, China\\
$^2$Department of Computer Science, Jinan University, Guangzhou, China\\
$^3$Institute for Integrated and Intelligent Systems, Griffith University, Brisbane, Australia\\
$^4$Department of Computer and Information Sciences, Temple University, Philadelphia, USA\\
$^5$Department of Computer Science, Sun Yat-sen University, Guangzhou, China\\
yifan.sysu@gmail.com;
\{nan.li, latecki\}@temple.edu;
k.su@griffith.edu.au
}

\begin{document}

\maketitle

\begin{abstract}
The tabu and restart are two fundamental strategies for local search. In this paper, we improve the local search algorithms for solving the Maximum Weight Clique (MWC) problem by introducing new tabu and restart strategies. Both the tabu and restart strategies proposed are based on the notion of a local search scenario, which involves not only a candidate solution but also the tabu status and unlocking relationship.  Compared to the strategy of configuration checking,  our tabu mechanism discourages forming a cycle of unlocking operations.  Our new restart strategy is based on the re-occurrence of a local search scenario instead of that of a candidate solution. 
Experimental results show that the resulting MWC solver outperforms several state-of-the-art solvers on the DIMACS, BHOSLIB, and two benchmarks from practical applications.
\end{abstract}

\section{Introduction}\label{sec:Introduction}

The maximum weight clique (MWC) problem is defined on a simple undirected graph $G = (V, E, w)$ where $V$ is the vertex set, an edge $e\in E$ is a 2-element subset of $V$, and $w:V\mapsto \Rz$ is a weighting function on $V$.
A \emph{clique} $C$ is a subset of $V$ such that each pair of vertices in $C$ is mutually adjacent.
The MWC problem is to find a clique with the greatest total weight.
This problem exists in many real-world applications like \cite{DBLP:conf/nips/BrendelT10,DBLP:conf/cvpr/BrendelAT11,DBLP:conf/nips/LiL12}.

Currently there are two types of algorithms for solving the MWC problem: complete ones \cite{yamaguchi2008new,shimizu2012some,DBLP:journals/jair/FangLX16,DBLP:conf/aaai/JiangLM17} and incomplete ones \cite{DBLP:journals/heuristics/Pullan08,DBLP:journals/anor/WuHG12,DBLP:conf/ijcai/CaiL16,DBLP:journals/eor/ZhouHG17,Nogueira2017}. The incomplete algorithms are designed to find a ``good" clique within reasonable time periods. In this paper, our focus is on local search, a widely accepted approach for the incomplete MWC algorithms. 
\subsection{Tabu and Restart in Local Search for MWC}
Local search, however, often suffers from the cycling problem, $\ie$, a candidate solution may be visited repeatedly. 
To deal with the cycling problem, we may adopt the tabu strategy \cite{DBLP:journals/anor/WuHG12}.  
The idea is that if the search flips a vertex's state ($\ie$, puts the vertex into current candidate solution or moves it out), then the vertex should be forbidden to return to its previous state for a certain period of search steps.   
A crucial issue here is when to relieve such a forbidding or tabu on the vertex.   
Configuration Checking (CC) is an effective strategy to resolve this issue and has been widely used in state-of-the-art MWC solvers \cite{DBLP:conf/aaai/WangCY16,DBLP:conf/ijcai/FanLLMLS17,DBLP:conf/ictai/FanMKCRLL17}. 
The idea of CC is that the tabu on a vertex may be relieved if one of its neighbors is flipped. 
In this case, we say that the flipped neighbor {\it unlocks} the vertex.    
Note that the unlocking operations among vertices may form a small cycle, which may lead a CC-based local search being stuck in a cycle. To escape from such a cycle, a CC-based local search usually needs some other diversifying strategies like constraint weighting \cite{DBLP:reference/crc/HoosS07a}, which is time-consuming and impractical for large and dense graphs.   
This calls for a new tabu strategy which discourages the unlocking cycles and allows the local search to move in a greater area.

Restart is another strategy \cite{battiti2001reactive} for resolving the cycling problem. Recently \cite{DBLP:conf/ijcai/FanLLMLS17} proposed a revisiting based restart strategy.
They proposed the notion of first growing step and set the triggering condition for a restart as revisiting a candidate solution at the first growing step.   
However, the local search with this restart triggering condition may restart too early to explore intensively.  Therefore,  we need to strengthen the triggering condition. To do so, we propose the notion of local search scenario, which involves not only the current candidate solution but also the tabu status and the unlocking relationship. 
Intuitively, when the current search revisits a candidate solution, it may be differentiated in the next step due to the different tabu status from before, and it does not need a restart. Moreover, when the current search revisits a candidate solution with the same tabu status, its tabu status may be differentiated in the next step due to the different unlocking relationship, and it does not need a restart either. Thus, to avoid restarting too early, the triggering condition should be based on the re-occurrence of a local search scenario.   

\subsection{Our Contributions}
As discussed above, we propose  new tabu and restart strategies based on the notion of a local search scenario.    
By using the tabu and the restart strategies, we develop an MWC solver named TRSC (Tabu and Restart with Scenario Checking).  
Similar to the CC strategy, the proposed tabu mechanism may relieve the tabu on a vertex by flipping a neighbor of the vertex, $\ie$,  the vertex can be unlocked by its neighbor, but this cannot be done by the same neighbor twice in a row. 
Our new restart strategy is based on the re-occurrence of a local search scenario instead of that of a candidate solution; in other words,  if a candidate solution is revisited together with the same tabu status and unlocking relationships as before, the search needs a restart.

We are the first to use the notion of local search scenario for both the tabu and restart strategies. For implementing a tabu strategy,  this work maintains a local search scenario, while previous approaches do not consider the unlocking relationship. For the restart purpose,  this work computes the hash value of local search scenarios, while previous approaches do the hash of visited candidate solutions.  Moreover,
our tabu and restart strategy interact and cooperate well. Since we employ a tabu strategy which is more restrictive than the strong configuration checking (SCC) strategy \cite{DBLP:conf/aaai/WangCY16}, our local search can travel in a loop bigger than before. So if we still use the previous restart strategies like those in \cite{DBLP:conf/ijcai/FanLLMLS17} and \cite{DBLP:conf/ictai/FanMKCRLL17}, the search will restart before a loop is visited completely, which we think is too early. 

To show the effectiveness of our approach, we compare our solver with state-of-the-art ones:  LSCC \cite{DBLP:conf/aaai/WangCY16}, RRWL \cite{DBLP:conf/ijcai/FanLLMLS17} and TSM-MWC \cite{JiangLLM18} on the DIMACS \cite{Johnson:1996:CCS:548182} and BHOSLIB \cite{DBLP:conf/ijcai/XuBHL05} benchmarks\footnote{http://sites.nlsde.buaa.edu.cn/$\sim$kexu/benchmarks/graph-benchmarks.htm}\footnote{https://github.com/notbad/RB/tree/master/generator/instances}, which were used in a wide range of recent papers.
We also compare these solvers on some graphs from real-world applications, $\ie$, the Winner Determination Problem (WDP) \cite{DBLP:conf/sigecom/Leyton-BrownPS00,DBLP:conf/ictai/LauG02,DBLP:journals/ai/Sandholm02}\footnote{http://www.info.univ-angers.fr/pub/hao/WDP/WDPinstance.rar}, the Error-correcting Codes (ECC) \cite{Ostergard:2001:NAM:766502.766504}, the Kidney-exchange Schemes (KES) and the Research Excellence Framework (REF) \cite{DBLP:conf/cp/McCreeshPST17}\footnote{https://github.com/jamestrimble/max-weight-clique-instances}.
Experimental results show that our solver outperforms several state-of-the-art solvers on the DIMACS, BHOSLIB, and some benchmarks from practical applications.
Furthermore, it is comparable with state-of-the-art on the remaining benchmarks.

\section{Preliminaries}
 We say that $u$ and $v$ are neighbors,  or $u$ and $v$ are adjacent to each other, if there is an edge $e = \{u, v\}$.
Also we use $N(v)$ to denote $\{u | u\textrm{ and }v\textrm{ are neighbors.}\}$, the set of $v$'s neighbors.
A maximal clique is a clique which is not a subset of any other clique.
Given a weighting function $w:V\mapsto \Rz$, the weight of a clique $C$, denoted by $w(C)$, is defined to be $\sum_{v\in C}w(v)$.
We use $age(v)$ to denote the number of steps since last time $v$ changed its state (inside or outside the candidate clique).
Given two vertices $v_i$ and $v_j$ where $i, j \in N^+$, we say $v_i < v_j$ if $i < j$.
Let $e$ be a bijection $e: E \leftrightarrow N$, which gives each edge an integer id between $0$ and $|E|-1$. Therefore, given two vertices $u$ and $v$,  $e(\{u, v\})$ denotes the id of the edge which connects $u$ and $v$.

\subsection{The Benchmark}
As to the DIMACS and the BHOSLIB benchmarks, we first obtain the Maximum Clique instances or convert the Maximum Independent Set instances into the complement graphs.
Then we use the method in \cite{DBLP:journals/heuristics/Pullan08} to generate the vertex weights, $\ie$,
for the $i$-th vertex $v_i$, $w(v_i)=(i \mod 200) + 1$.
Also, we compare state-of-the-art MWC solvers on a list of  benchmarks from practical applications.
\subsection{Multi-neighborhood Search}
In order to find a good clique, the local search usually moves from one clique to another until the cutoff arrives, then it returns the best clique that has been found.
There are three operators: $\add$, $\swap$ and $\drop$, which guide the local search to move in the clique space.
In \cite{DBLP:conf/ausai/FanLMWSS16} two sets were defined as below which ensures that the clique property is preserved: 
\begin{displaymath}
S_{add}(C) = \left\{ \begin{array}{ll}
\{v | v \not\in C, v \in N(u)\textrm{ for all }u \in C\} & \textrm{if $|C|>0$;}\\
\emptyset & \textrm{if $|C|=0$.}
\end{array} \right.
\end{displaymath}

\begin{displaymath}
S_{swap}(C) = \left\{ \begin{array}{ll}
\{(u, v) | u \in C, v \not\in C, \{u, v\} \not\in E, \\v \in N(w)\textrm{ for all } w \in C \backslash \{u\}\} & \textrm{if $|C|>1$;}\\
\emptyset & \textrm{if $|C|\leq 1$.}
\end{array} \right.
\end{displaymath}

For simplicity we will write $S_{add}$ and $S_{swap}$ in short for $S_{add}(C)$ and $S_{swap}(C)$ respectively.
We use $\Delta_{add}$, $\Delta_{swap}$ and $\Delta_{drop}$ to denote the increase of $w(C)$ for the operations $\add$, $\swap$ and $\drop$ respectively.
Obviously, we have (1) for a vertex $v \in S_{add}$, $\Delta_{add}(v) = w(v)$; (2) for a vertex $u \in C$, $\Delta_{drop}(u) = -w(u)$; (3) for a vertex pair $(u, v) \in S_{swap}$, $\Delta_{swap}(u, v) = w(v) - w(u)$.

\subsection{The Strong Configuration Checking Strategy}\label{sec:scc}
Recently, \cite{DBLP:journals/ai/CaiSS11} proposed the configuration checking (CC) strategy to reduce cycling.
The CC strategy works as follows.
If a vertex is removed out of the candidate set, it is forbidden to be added back into the candidate set until its configuration has been changed. 
Typically, the configuration of a vertex refers to the state of its neighboring vertices. 

The CC strategy is usually implemented with a Boolean array named $\mathit{confChange}$, where $\mathit{confChange}(v)=1$ means that $v$'s configuration has changed since last time it was removed, and $\mathit{confChange}(v)=0$ otherwise. 

Later \cite{DBLP:conf/aaai/WangCY16} modified CC into a more restrictive version, which is called strong configuration checking (SCC), to deal with the MWC problem. 
The main idea of the SCC strategy is as follows: after a vertex $v$ is dropped from or swapped from $C$, it can be added or swapped back into $C$ only if one of its neighbors is added into $C$.

In details, the SCC strategy works as follows.
(1) Initially $\confChange(v)$ is set to 1 for each vertex $v$;
(2) When $v$ is added, $\confChange(n)$ is set to 1 for all $n \in N(v)$;
(3) When $v$ is dropped, $\confChange(v)$ is set to 0;
(4) When $(u, v)\in S_{swap}$ are swapped, $\confChange(u)$ is set to 0.
Lastly $\confChange(v)$ is also referred to as $v$'s tabu status. 

\subsection{A Fast Hashing Function}\label{sec:fans-hash-func-17}
\cite{DBLP:conf/ijcai/FanLLMLS17} proposed a fast hashing function as below which detects revisiting both efficiently and effectively.

\begin{definition}\label{hash-func}
Given a clique $C$ and a prime number $p$, we define the hash value of $C$, denoted by $\hash(C)$, as $(\sum_{v_i \in C}{2^i}) \mod p$, which maps a clique $C$ to its hash entry $hash(C)$.
\end{definition}


At the beginning, they calculate $(2^i \mod p)$ iteratively with different values of $i$, based on the proposition below.
\begin{prop}
 $2^i \mod p = 2(2^{i-1}\mod p) \mod p$.
\end{prop}

These values are then saved in an array for later references.
Hence, in Theorem \ref{hash-update} below, the subformulas $(2^i \mod p)$ can be computed in constant complexity.
So the hash value of the current clique can be updated in $O(1)$ complexity as well.

\begin{theorem}\label{hash-update}
Let $C$ be the current clique, then we have 
\begin{enumerate}
\item $\hash(C\cup\{v_i\}) = [\hash(C) + (2^i \mod p)] \mod p$;
\item $\hash(C\backslash\{v_i\})\! =\! [\hash(C) + p - (2^i\! \mod p)]\! \mod p$.
\end{enumerate}
\end{theorem}


\subsection{Review of LSCC}
LSCC consists of two procedures: randomly generating a maximal clique $C$ and improving $C$ in a deterministic way.
In each local move, LSCC selects the neighboring clique with the greatest weight according to the SCC criterion. 
Every $4,000$ steps, the search is restarted.

\cite{DBLP:conf/ictai/FanMKCRLL17} showed that without restarts, LSCC may fall into a dead loop, $i.e.$, no matter how many steps it performs, it always miss the optimal solution (See Example \ref{example:dead-loop}).
Here we cite their example graph. 
In the next section, we will explain why LSCC is misled and propose a new tabu strategy to deal with this case.

\begin{example}\label{example:dead-loop}
Consider the graph $G$, where $w(v_i) = i\cdot 10$ for any $i \neq 3$ and $w(v_3) = 3$. 
Obviously the optimal solution in $G$ is $\{v_3, v_5, v_6, v_8\}$.

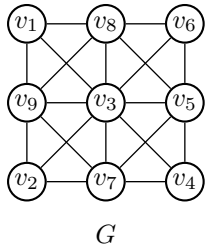
    \begin{center}
    \begin{tikzpicture}[thick, scale=0.7]
    \tikzstyle{every node}=[minimum size=5mm,inner sep=0pt,draw, fill=white]
    \path
      (0, -1.5) node(1)[circle] {$v_3$}
      (-1.5, 0) node(2)[circle] {$v_1$}
      (-1.5, -3) node(3)[circle] {$v_2$}
      (1.5, -3) node(4)[circle] {$v_4$}
      (1.5, 0) node(5)[circle] {$v_6$}
      (0, -3) node(6)[circle] {$v_7$}
      (1.5, -1.5) node(7)[circle] {$v_5$}
      (0, 0) node(8)[circle] {$v_8$}
      (-1.5, -1.5) node(9)[circle] {$v_9$};
      \node[draw=none,fill=none] at (0, -4) {$G$};
  \foreach \source/\target in {1/4, 1/2, 3/1, 1/5, 1/8, 1/9, 9/8, 4/6, 4/7, 6/7, 1/7, 7/8, 2/8, 5/8, 3/9, 3/6, 2/9, 5/7, 6/9, 1/6}
  \draw[-,black,line width=0.5pt] (\source) --(\target);
\end{tikzpicture}
    \end{center}
\begin{enumerate}
\item Initially at Step 1, $C = \emptyset$. 
Suppose we select $v_2$ as the first vertex and put it into $C$, then LSCC obtains a clique $C = \{v_2, v_3, v_7, v_9\}$. 
Meanwhile $\confChange(v) = 1$ for all $v \in V$.

\item Next the local search reaches $\{v_1, v_3, v_8, v_9\}$ at Step 9. At the same time $\confChange(v)=1$ for all $v \in V$.

\item Then the local search moves back to $\{v_2, v_3, v_7, v_9\}$ at Step 14. Meanwhile $\confChange(v)=1$ for all $v \in V$.

\end{enumerate}
Then the local search repeats the steps above and is restricted in a cycle, without finding the optimal solution.
\end{example}


\section{Tabu and Restart with Scenario Checking}
We propose a tabu strategy and a restart strategy based on the notion of a local search scenario. The two strategies coordinate with each other.
\subsection{Forbidding Repeated Unlocking}
We say that vertex $u$ is unlocked by its neighbor $v$ if the tabu on $u$ is relieved just after $v$ is flipped.   Note that a vertex can be unlocked by different vertices in different steps. We use $\unlocker(v)$ to denote the last vertex which unlocks $v$.
We use $U$ to denote the unlocking relation, $\ie$, $U =\{(v_1, v_2) | \unlocker(v_1) = v_2\textrm{ and }v_1, v_2 \in V\}$.
Therefore given $(u_1, u_2) \in U$, it can be read as $u_1$ was unlocked by $u_2$ last time.
\subsubsection{Unlocking Graph}

Based on Example \ref{example:dead-loop}, we have a graph below, which is called \emph{unlocking graph}.
It describes the unlocking operations during the local search dead loop.
For instance, at Step 8 $v_1$ unlocks $v_3$, and at Step 9 $v_3$ unlocks $v_2$ and $v_7$.
Each time the search traverses the local search dead loop, each unlocking operation in the graph will be performed once.

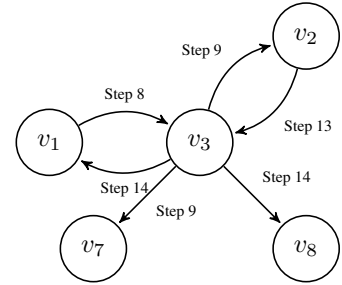
\begin{center}
\begin{tikzpicture}[->,>=stealth',shorten >=1pt,auto,node distance=2.0cm,semithick]
  \tikzstyle{every state}=[text=black]

  \node[state] (A)                    {$v_3$};
  \node[state]         (C) [below right of=A] {$v_8$};
  \node[state]         (B) [above right of=A] {$v_2$};
  \node[state]         (D) [below left of=A]  {$v_7$};
  \node[state]         (E) [left of=A]  {$v_1$};

  \path (A) edge [bend left]  node {\tiny Step 9} (B)
            edge [bend left]  node {\tiny Step 14} (E)
            edge              node {\tiny Step 9} (D)
            edge              node {\tiny Step 14} (C)

        (B) edge [bend left]  node {\tiny Step 13} (A)
        (E) edge [bend left]  node {\tiny Step 8} (A);
\end{tikzpicture}
\end{center}

In this unlocking graph, we observe that there are two unlocking cycles: $\{v_1, v_3\}$ and $\{v_2, v_3\}$.
These unlocking cycles and the best-picking heuristic together lead the search back to a visited solution, with the tabu status the same as before.
So the local search is restricted in a dead loop.
\subsubsection{Our Strategy}
In this situation, we need to delete some of the directed edges in the unlocking graph so that the local search can escape.
So we propose a tabu management strategy as below, which is called Forbidding Repeated Unlocking (FRU).

\begin{enumerate}
\item Initially $\free(v)\!\leftarrow\! 1$ and $\unlocker(v)\! \leftarrow\! \mathit{NULL}$ $\forall$$v \in V$;
\item When $v$ is added into $C$, 
\begin{enumerate}
\item $\free(v)$ is set to $1$,
\item $\forall$ $n \in N(v)$ s.t. $\free(n)$ $=$ $0$ and $\unlocker(n)$ $\neq$ $v$, $\free(n)\leftarrow 1$ and $\unlocker(n)\leftarrow v$;
\end{enumerate} 
\item When $v$ is dropped or swapped from $C$, $\free(v)\leftarrow 0$.
\end{enumerate}
Initially when all vertices are free, none of the vertices has been unlocked, so $\unlocker(v)$ is set to $\NULL$ for all $v \in V$.
In Item (b), $n$ can be unlocked by $v$ only if $n$ was not unlocked by $v$ last time, $\ie$, $n$ cannot be unlocked by $v$ twice in a row.
Item (a) is tricky and will be explained in Section \ref{sec:hash-restart}.
We use $F = \{v | \free(v) = 1\}$ denote the set of free vertices, and $\free(v)$ is also referred to as $v$'s tabu status.

\subsection{Considering Tabu in the Restart Strategy}\label{sec:hash-restart}
We use \emph{local search scenario} to describe the solution, the tabu status and the unlocking relation as a whole in a given step.
\begin{definition}
The local search scenario in Step $t$, denoted by $\mathcal{S}_t$, is defined as a tuple which consists of the solution $C_t$, the tabu status $F_t$ and the unlocking relation $U_t$ in Step $t$, $\ie$, $\mathcal{S}_t = \langle C_t, F_t, U_t \rangle$.
\end{definition}
So a local search scenario depicts much information which will determine the following local search steps to a great extent.
In other words, if a local search scenario re-occurs, the search may probably be restricted in a cycle.
For simplicity, we write a local search scenario as $\mathcal{S} = \langle C, F, U \rangle$.

\subsubsection{An Extended Hash Function}

We use a hash table to approximately detect the re-occurrence of a local search scenario.
Since the collisions, $\ie$, different scenarios may share the same hash entry, are rare in our settings, we do not resolve them.
Below we define a hash function where $p$ is a prime number.
\begin{eqnarray}\nonumber
hash(\mathcal{S})&\!\!\!\!\!\!\!\!\!\!\!\!\!\!\!\!\!\!\!\!=\!\!\!\!\!\!\!\!\!\!\!\!\!\!\!\!\!&\Big[\Sigma_{v_i \in C}2^i + \Sigma_{v_i \in F}2^{|V|+i} \\ \nonumber
&& \left. + \Sigma_{(v_i, v_j) \in U, i < j}2^{2|V| + 1 + e(\{v_i, v_j\})} \right.\\ \nonumber
&& \left. + \Sigma_{(v_i, v_j) \in U, i > j}2^{2|V| + |E| + 1 + e(\{v_i, v_j\})}\Big] \!\!\!\!\mod p. \right.
\end{eqnarray}

So far as we know, all previous hashing strategies compute the hash value of a candidate solution, $\eg$, \cite{battiti2001reactive}, and we are the first time to compute the hash value of a local search scenario.
During the search, we will use the methods in Section \ref{sec:fans-hash-func-17} to maintain the hash value of the current local search scenario, and we set $p = 10^9+7$.
With this prime number $p$, our hash table consumes around 1 GB memory.
In our experiments, our solver performs less than $10^7$ steps in any run.
Therefore given the $10^9+7$ hash entries, the number of collisions is negligible.

Now we return to Item (a) in our tabu rules.
In usual local search solvers, a vertex $u\in C$ is always allowed to be removed.
In this sense, whether $\free(u) = 1$ or $\free(u) = 0$   does not matter, hence, we always set $\free(u)$ to be $1$ so that this unimportant difference will not affect the hash value.

\section{The TRSC Algorithm}

The top level algorithm is shown in Algorithm \ref{TRSC}, where the \texttt{localMove()} procedure is shown in Algorithm \ref{algo:local-move}.

\begin{algorithm}
  \SetKwData{Left}{left}\SetKwData{This}{this}\SetKwData{Up}{up}
  \SetKwFunction{Union}{Union}\SetKwFunction{FindCompress}{FindCompress}
  \SetKwInOut{Input}{input}\SetKwInOut{Output}{output}

  \Input{A graph $G = (V, E, w)$ and the \emph{cutoff}}
  \Output{The best clique that was found}
  \BlankLine
  
  $C^* \leftarrow C \leftarrow \emptyset$;
   $step \leftarrow 1$;
  $\free(v) \leftarrow 1$ for all $v \in V$;\\
  \lWhile{elapsed time $<$ cutoff}{
  	localMove()
  }
  \Return $C^*$;
  \caption{TRSC}\label{TRSC}
\end{algorithm}


\begin{algorithm}[t]
  \SetKwData{Left}{left}\SetKwData{This}{this}\SetKwData{Up}{up}
  \SetKwFunction{Union}{Union}\SetKwFunction{FindCompress}{FindCompress}
  \SetKwInOut{Input}{input}\SetKwInOut{Output}{output}

	\If{$C = \emptyset$}{\label{rand-walk}
    	add a random vertex into $C$;\label{init-vertex}\\
        \lWhile{$S_{add} \neq \emptyset$}{\label{rand-construct}
        		add a random vertex from $S_{add}$
        	}
                    $\lastStepImproved \leftarrow $ \emph{true};
        }
    $v \leftarrow $ a vertex in $S_{add}$ such that $\mathit{free}(v)=1$ with the biggest $\Delta_{add}$; otherwise $v \leftarrow\textit{NL}$;\\
    $(u, u') \leftarrow$ a pair in $S_{swap}$ such that $\free(u')=1$ with the biggest $\Delta_{swap}$; otherwise $(u, u') \leftarrow \textit{(NL, NL)}$;\\
     \If{$v \neq \textit{NL}$}{\label{}   				
                    \textbf{if} $(u, u') = \textit{ (NL, NL) }$  \emph{or} $\Delta_{add} > \Delta_{swap}$
                    \textbf{then} $C \leftarrow C \cup \{v\}$;
                    \textbf{else} $C \leftarrow C\backslash\{u\} \cup \{u'\}$;\\
                    $\mathit{lastStepImproved} \leftarrow $ \emph{true};
                }\label{}
                \Else{
               	\If{$(u, u') = \textit{(NL, NL)}$ or $\Delta_{swap} < 0$}{\label{no-improvement}

                	\If{$\lastStepImproved = $ \emph{true}}{\label{last-step-improved}
                    					\lIf{$w(C) > w(C^*)$}{$C^* \leftarrow C$} 
                                       
                        \If{$hash(\mathcal{S})$ is marked}{\label{hash-check}
             drop all vertices in $C$; 
             $step$++;
             \Return;
                    }
                     mark $hash(\mathcal{S})$;\label{hash-mark}
                        }
                 $\lastStepImproved \leftarrow $ \emph{false};
                    }
                    \Else{
                    	$\lastStepImproved \leftarrow $ \emph{true};
                    }
                	$x \leftarrow $ a vertex in $C$ with the biggest $\Delta_{drop}$;\\
                     \textbf{if} $(u, u') = \textit{ (NL, NL) }$  \emph{or} $\Delta_{drop} > \Delta_{swap}$
                    \textbf{then} $C \leftarrow C \backslash \{x\}$;
                    \textbf{else} $C \leftarrow C\backslash\{u\} \cup \{u'\}$;\\

                        }\label{}
                    apply FRU rules; $step$++;

  \caption{localMove}\label{algo:local-move}
\end{algorithm}

In Algorithm \ref{algo:local-move}, the arguments of the functions are explicit from the context and thus omitted.
All ties are broken in favor of the oldest one just like LSCC.
We employ a predicate $\lastStepImproved$ s.t. $\lastStepImproved$ $= $ \emph{true} iff the clique weight was increased in the last step.
Then we use $lastStepImproved$ to identify local optima.
When both the conditions in Lines \ref{no-improvement} and \ref{last-step-improved} hold, a local optimum is reached.
We will mark and detect the occurrence of a local search scenario only at local optima, because we desire to decrease the number of hash entries that need to be marked.
So the hash collisions hardly exist.

Now we run TRSC on the graph $G$ in Example \ref{example:dead-loop} as below.
\begin{example}\label{example:lstm-run}
\begin{enumerate}
\item Initially $C = \emptyset$. 
Like Example \ref{example:dead-loop}, suppose we select $v_2$ as the first vertex, then TRSC obtains a clique $C = \{v_2, v_3, v_7, v_9\}$ which is a local optimum.  
At this time $S_{add} = S_{swap} = \emptyset$. 
Meanwhile $\free(v) = 1$ and $\unlocker(v) = \NULL$ for all $v \in V$, so $U = \emptyset$.
We denote this local search scenario by $\mathcal{S}^1$.
\item Then we will perform steps just like those in Example \ref{example:dead-loop}.
More specifically the local search moves to $\{v_1, v_3$, $v_8, v_9\}$ which is also a local optimum, and we denote this local search scenario as $\mathcal{S}^2$. 
Then the local search moves back to $\{v_2, v_3, v_7, v_9\}$.
At this time, $\free(v) = 1$ for all $v \in V$, and $U = \{(v_1, v_3)$, $(v_2, v_3)$, $(v_7, v_3)$, $(v_8, v_3), (v_3, v_2)\}$.
We denote this local search scenario by $\mathcal{S}^3$.
Notice that not only has the solution $C$ been revisited in this step, the tabu status $F$ has also become the same as before. 
However, the unlocking relation $U$ is not the same as before, so $\mathcal{S}^1 \neq \mathcal{S}^3$, $\ie$, this current local search scenario has not occurred before.
Hence, our solver does not restart.
\item Next the local search moves to $\{v_1, v_3, v_8, v_9\}$ again, and the last vertex which enters $C$ is $v_3$.
However, things are different at this time because of the FRU strategy.
Since $v_2$ and $v_7$ was unlocked by $v_3$ last time,  neither of them can be unlocked by $v_3$ this time.
That is, $\free(v_2) = \free(v_7) = 0$ still holds at this time, which will \emph{prevent} the local search from moving back to $\{v_2, v_3, v_7, v_9\}$.
We denote this current local search scenario by $\mathcal{S}^4$.
Notice that $\mathcal{S}^2\neq \mathcal{S}^4$.
\item So the local search changes its direction and move to $\{v_3, v_5, v_6, v_8\}$ which is the optimal solution.
\end{enumerate}
So we see that TRSC can search a local area more thoroughly than LSCC.
Notice that if we adopted previous restart strategies like those in \cite{DBLP:conf/ijcai/FanLLMLS17} and \cite{DBLP:conf/ictai/FanMKCRLL17}, the search would restart before the occurrence of $\mathcal{S}^4$.
In this case the FRU strategy behaves simply the same as the SCC strategy.
\end{example}

\section{Implementations}
In this section we will show how to implement the tabu and the restart strategy when a vertex is added (See Algorithm \ref{algo:add-v}).

In Algorithm \ref{algo:add-v}, Lines \ref{algo-add-update-hash-wrt-add}, \ref{algo-add-update-hash-wrt-unlock-itself}, \ref{algo-add-update-hash-wrt-check-unlock}, \ref{algo-add-update-hash-wrt-delete}, \ref{algo-add-update-hash-wrt-insert} and \ref{algo-add-update-hash-wrt-unlock-nb} show how to update the hash value of the current local search scenario,
while the other lines implement our proposed tabu strategy.
Notice Line \ref{algo-add-update-hash-wrt-check-unlock}.
If a vertex is unlocked for the first time, then no vertices have ever unlocked it, so we do not delete the respective tuple.


In the Algorithm \ref{algo:add-v}, our solver  will need to compute the value of $e(\{n, \unlocker(n)\})$.
It will do this just as what CERS\footnote{https://github.com/Fan-Yi/Local-Search-for-Maximum-Edge-Weight-Clique} \cite{DBLP:conf/ictai/FanMKCRLL17} does when solving the maximum edge weight clique problem, so we have 
\begin{prop}
Computing $e(\{n, \unlocker(n)\})$ in Algorithm \ref{algo:add-v} can be done in $O(1)$ complexity.
\end{prop}
Considering that $(2^i \mod p)$ for $1\leq i \leq 2|V|+2|E|$ has been computed and stored in an array before, we have 
\begin{prop}
In any local search step, \\$(2^{|V|+i}\mod p)$, $(2^{2|V| + 1 + e(\{n, \unlocker(n)\})} \mod p)$ and $(2^{2|V| + |E| + 1 + e(\{n, \unlocker(n)\})} \mod p)$ can all be computed in $O(1)$ complexity.
\end{prop}
So all the lines which update the hash value of a local search scenario can be executed in $O(1)$ complexity.
For example Line \ref{algo-add-update-hash-wrt-delete} can be implemented as follows.
$hash(\mathcal{S}) \leftarrow (hash(\mathcal{S}) + p - 2^{2|V| + 1 + e(\{n, \unlocker(n)\})} \mod p) \mod p$ if $n < \unlocker(n)$, and  $hash(\mathcal{S}) \leftarrow (hash(\mathcal{S}) + p - 2^{2|V| + |E| + 1 + e(\{n, \unlocker(n)\})} \mod p) \mod p$ otherwise.
On the other hand, Line \ref{algo-add-update-hash-wrt-insert} can be implemented as follows.
$hash(\mathcal{S}) \leftarrow (hash(\mathcal{S}) + 2^{2|V| + 1 + e(\{n, \unlocker(n)\})} \mod p) \mod p$ if $n < \unlocker(n)$, and  $hash(\mathcal{S}) \leftarrow (hash(\mathcal{S}) + 2^{2|V| + |E| + 1 + e(\{n, \unlocker(n)\})} \mod p) \mod p$ otherwise.
Therefore we have 
\begin{theorem}
\begin{enumerate}
\item The complexity of maintaining the hash value of a local search scenario wrt to \texttt{add$(v)$} and \texttt{drop$(v)$} is $O(d(v))$, where $d(v)$ is the degree of $v$.
\item The respective complexity for \texttt{swap$(u, v)$} is $O(d(u)+d(v))$.
\end{enumerate}

\end{theorem}

\begin{algorithm}
  \SetKwData{Left}{left}\SetKwData{This}{this}\SetKwData{Up}{up}
  \SetKwFunction{Union}{Union}\SetKwFunction{FindCompress}{FindCompress}
  \SetKwInOut{Input}{input}\SetKwInOut{Output}{output}

  update $hash(\mathcal{S})$ wrt. add $v$ into $C$;\label{algo-add-update-hash-wrt-add}\\
  \If{$\free(v)=0$}{
    	$\free(v) \leftarrow 1$;\\
        update $hash(\mathcal{S})$ wrt. unlock $v$;\label{algo-add-update-hash-wrt-unlock-itself}
    }
    \ForEach{$n \in N(v)$}{
  		\lIf{$\free(n)=1$ or $\unlocker(n) = v$}{
        	continue
    }
    \If{$\unlocker(n) \neq NULL$}{\label{algo-add-update-hash-wrt-check-unlock}
        update $hash(\mathcal{S})$ wrt. delete $(n, \unlocker(n))$;\label{algo-add-update-hash-wrt-delete}
    }
    $\free(n) \leftarrow 1$; $\unlocker(n) \leftarrow v$;\\
    update $hash(\mathcal{S})$ wrt. insert $(n, \unlocker(n))$;\label{algo-add-update-hash-wrt-insert}\\
    update $hash(\mathcal{S})$ wrt. unlock $n$;\label{algo-add-update-hash-wrt-unlock-nb}
    
  }
  \caption{add($v$)}\label{algo:add-v}
\end{algorithm}

\section{Empirical Evaluations}
We compare our solver to state-of-the-art complete and incomplete solvers including TSM-MWC, RRWL and LSCC.
\subsection{Experimental Protocol}
For LSCC, the search depth $L$ was set to 4,000 as is in \cite{DBLP:conf/aaai/WangCY16}.
TSM-MWC was compiled by gcc 6.3.0 with -O3 option and all other solvers were compiled by g++ 4.7.3 with -O3 option.
The experiments were conducted on a cluster equipped with Intel(R) Xeon(R) CPUs
X5650 @2.67GHz with 16GB RAM, running Red Hat Santiago OS.
Since TSM-MWC is an exact solver, it was executed on each instance only once.
Each other solver was executed on each instance with seeds from 1 to 100. 
The cutoff was set to 3,600s for each solver on each instance.

In each table, we report the maximum weight (``$w_{max}$") and averaged weight (``$w_{avg}$") of the cliques found by the algorithms.
As to TSM-MWC, if it is able to confirm the optimality of the returned solution, we mark * in the respective table entry; otherwise, we report that best-found solution within the cutoff. 
Also since TSM-MWC was executed on each instance only once, we used the weight of the returned solution as both the $w_{max}$ and the $w_{avg}$ values.
In each table, we only list those graphs on which all solvers did not find the same $w_{max}$ or $w_{avg}$ values.
For the sake of space, we abbreviate some instance names.

\subsection{The Benchmarks}
We considered two types of datasets: (1) the DIMACS and the BHOSLIB benchmarks; (2) the benchmarks from practical applications including Winner Determination Problem (WDP), Error-correcting Codes (ECC), Research Excellence Framework (REF) and Kidney-exchange Schemes (KES).

In details, the WDP instances are divided into three test sets.
(1) The first set contains 499 \textbf{relatively easy} instances provided by \cite{DBLP:conf/ictai/LauG02} with up to 1500 items and 1500 bids. 
These instances are divided into 5 different groups, each group labeled as REL-$m$-$n$, where $m$ is the number of items and $n$ is the number of bids. 
(2) The second set contains 20 \textbf{challenging} instances obtained from a generator provided by
\cite{DBLP:journals/ai/Sandholm02} (SAND).
(3) The third set contains 10 \textbf{challenging} instances generated randomly by the program combinatorial auction test suite (CATS) generator developed by \cite{DBLP:conf/sigecom/Leyton-BrownPS00}. 
Furthermore the REF instances are divided into two test sets.
(1) The first one (GOLD-RAG-REF) contains 100 \textbf{relatively easy} instances. 
(2) The second one (RAG-REF) contains 29 \textbf{challenging} instances.
Lastly the KES dataset contains 100 instances named from \texttt{001.wclq} to \texttt{100.wclq}. We only used those 50 \textbf{challenging} instances named from \texttt{051.wclq} to \texttt{100.wclq}, since the first 50 instances are easy to solve.
Lastly the ECC benchmark contains 15 instances which are all \textbf{relatively easy}.

In each group of the \textbf{relatively easy} instances, the solutions returned by TRSC were all proved to be optimal by TSM-MWC, $\ie$, it found the optimal solution in any run.
In Table \ref{tab:easy-graphs}, we present the averaged time (seconds) needed to locate the respective solutions for each solver in each group (``LocateTime").
Since TSM-MWC is able to confirm the optimality of the returned solution, we also report the time needed to find and prove the optimal solution (``ConfirmTime'').

\begin{table} [t]\scriptsize
\setlength{\tabcolsep}{8.0pt}
\renewcommand{\arraystretch}{1.00}
	\centering
	\caption{Experimental Results on Relatively Easy Graphs}
    \begin{tabular}{| l |l | l | l | }
    \hline
    Graph &  \multicolumn{2}{|c|}{TSM-MWC}& TRSC  \\
    \cline{2-4}
     &LocateTime &ConfirmTime &LocateTime \\ \hline \hline
    REL-500-1000 &18.53&23.86&27.37 \\
    REL-1000-1000 &0.7&0.95&3.06 \\
    REL-1000-500 &0.02&0.03&0.23 \\
    REL-1000-1500 &0.57&0.8&3.54 \\
    REL-1500-1500 &0.98&1.32&3.99 \\ \hline
    ECC &4.34&13.65&0.07 \\ \hline
    GOLD-RAG-REF &0.02&1.59&0.01 \\

    \hline
    \end{tabular}\label{tab:easy-graphs}
    
\end{table}

\begin{table} [h]\scriptsize
\setlength{\tabcolsep}{1pt}
\renewcommand{\arraystretch}{1.25}
	\caption{Results on Vertex-weighted BHOSLIB and DIMACS}
    \centering
\begin{tabular}{|l|l|l|l|l|}
\hline
Graph &TSM-MWC &LSCC &RRWL &TRSC \\
	& $w$& $w_{max} (w_{avg})$& $w_{max} (w_{avg})$& $w_{max} (w_{avg})$\\
\hline
\hline
frb56-25-1 &3693 &5886(5834.58)  &5916(5841.13)  &5916(\textbf{5850.63}) \\
frb56-25-2 &4470 &5886(5826.08)  &5886(5827.72)  &5882(\textbf{5842.6}) \\
frb56-25-3 &3958 &5844(5792.07)  &5842(5795.35)  &\textbf{5854}(\textbf{5805.8}) \\
frb56-25-4 &4609 &5873(5833.78)  &5877(5830.09)  &5877(\textbf{5840.62}) \\
frb56-25-5 &4023 &5817(5766.64)  &5810(5774.23)  &\textbf{5843}(\textbf{5785.52}) \\
frb59-26-1 &4469 &6591(6548.68)  &6591(6539.59)  &6591(\textbf{6554.16}) \\
frb59-26-2 &5105 &6645(6558.62)  &6645(6552.96)  &6645(\textbf{6568.67}) \\
frb59-26-3 &4373 &6576(6523.49)  &6606(6532.8)  &6606(\textbf{6542.24}) \\
frb59-26-4 &4916 &6592(6501.58)  &6592(6505.26)  &6592(\textbf{6518.74}) \\
frb59-26-5 &5038 &\textbf{6584}(6527.69)  &6569(6523.45)  &6581(\textbf{6533.69}) \\
\hline
frb65-28-1 &5208 &7410(7319.63)  &7405(7353.08)  &\textbf{7432}(\textbf{7377.73}) \\
frb65-28-2 &4788 &7421(7369.89)  &7425(7365.99)  &\textbf{7441}(\textbf{7380.56}) \\
frb65-28-3 &4857 &\textbf{7449}(7359.52)  &7434(7361.07)  &7445(\textbf{7377.62}) \\
frb65-28-4 &4587 &7433(7366.07)  &7438(7366.83)  &\textbf{7448}(\textbf{7381.88}) \\
frb65-28-5 &4881 &7451(7354.89)  &7451(7354.73)  &7451(\textbf{7374.99}) \\
frb70-30-1 &5125 &7717(7597.72)  &7688(7600.97)  &\textbf{7772}(\textbf{7617.76}) \\
frb70-30-2 &5159 &7749(7668.98)  &7734(7667.96)  &\textbf{7777}(\textbf{7689.7}) \\
frb70-30-3 &4635 &7678(7622.72)  &\textbf{7733}(7620.69)  &7696(\textbf{7638.95}) \\
frb70-30-4 &4918 &7739(7680.75)  &7750(7684.43)  &\textbf{7766}(\textbf{7703.22}) \\
frb70-30-5 &5402 &\textbf{7749}(7669.0)  &7725(7666.93)  &7740(\textbf{7683.99}) \\
frb75-32-1 &5168 &8615(8537.8)  &\textbf{8637}(8532.98)  &8621(\textbf{8556.81}) \\
frb75-32-2 &5820 &8644(8569.07)  &8657(8570.63)  &\textbf{8667}(\textbf{8583.37}) \\
frb75-32-3 &5928 &\textbf{8619}(8506.95)  &8565(8496.14)  &8600(\textbf{8528.81}) \\
frb75-32-4 &6315 &\textbf{8714}(8589.85)  &8688(8590.29)  &8692(\textbf{8607.74}) \\
frb75-32-5 &5393 &\textbf{8663}(8590.24)  &8655(8591.12)  &8644(\textbf{8602.52}) \\
frb80-33-1 &5819 &9353(9249.81)  &\textbf{9461}(9242.11)  &9407(\textbf{9294.27}) \\
frb80-33-2 &5783 &9343(9251.12)  &\textbf{9390}(9248.75)  &9387(\textbf{9293.51}) \\
frb80-33-3 &5954 &\textbf{7449}(7359.52)  &7434(7361.07)  &7445(\textbf{7377.62}) \\
frb80-33-4 &6655 &9387(9305.08)  &\textbf{9430}(9301.91)  &9408(\textbf{9332.44}) \\
frb80-33-5 &6167 &9413(9337.37)  &9415(9341.21)  &\textbf{9478}(\textbf{9360.45}) \\
\hline
C2000.9 &8338 &10999(10942.27)  &10999(10951.67)  &10999(\textbf{10965.43}) \\
C4000.5 &2438 &2792(2792.0)  &2792(2792.0)  &2792(2792.0) \\
\hline
hamming10-4 &4828 &5129(5129.0)  &5129(5129.0)  &5129(5129.0) \\
\hline
keller6 &4793 &8062(7841.39)  &8062(7891.05)  &8062(\textbf{7961.9}) \\
\hline
MANN\_a45 &\textbf{34265}$^*$ &34254(34244.51)  &34263(34254.75)  &34262(34254.18) \\
MANN\_a81 &110037 &111126(111093.81)  &111346(\textbf{111306.32})  &\textbf{111356}(111305.79) \\
	\hline
	\end{tabular}\label{tab:vertex-weighted-BHOSLIB-DIMACS}
\end{table}
\vspace*{-11mm}

\begin{table} [h]\tiny
\setlength{\tabcolsep}{3.3pt}
\renewcommand{\arraystretch}{1.25}
	\caption{Results on SAND and CATS against Incomplete Solvers}
    \centering
\begin{tabular}{|l|l|l|l|}
\hline
Graph  &LSCC &RRWL &TRSC \\
	& $w_{max} (w_{avg})$& $w_{max} (w_{avg})$& $w_{max} (w_{avg})$\\
\hline
\hline
Decay\_100  &84535100(83250072.0)  &85281200(84059567.0)  &\textbf{85510000}(\textbf{84706031.0}) \\
Decay\_200  &149346900(145222082.0)  &151127500(146679501.0)  &\textbf{151727300}(\textbf{147776429.0}) \\
Decay\_300 &210276200(206690975.0)  &212143900(207974912.0)  &\textbf{214579300}(\textbf{208712490.0}) \\
Decay\_400 &255183300(249843205.0)  &257921100(251010990.0)  &\textbf{258063500}(\textbf{251908678.0}) \\
Decay\_500 &313889900(309658340.0)  &316621400(311568688.0)  &\textbf{321463500}(\textbf{311769095.0}) \\
\hline
uni\_500\_10  &26564200(26555349.0)  &26564200(\textbf{26558697.0})  &26564200(26553017.0) \\
\hline
arbi\_40  &40460122(39971566.63)  &40460122(40460122.0)  &40460122(40460122.0) \\
arbi\_100  &88001917(85989964.79)  &89310479(87487330.5)  &\textbf{89772104}(\textbf{87617454.9}) \\
\hline
matc\_80  &5083256(\textbf{5075016.11})  &5083256(5072696.82)  &5081870(5070041.87) \\
\hline
paths\_40  &\textbf{248514}(\textbf{247878.71})  &248236(247799.52)  &248236(247778.48) \\
paths\_100  &363213(360159.01)  &363449(361218.25)  &\textbf{364972}(\textbf{361397.42}) \\
\hline
regi\_40  &45588953(45430691.25)  &45588953(\textbf{45588953.0})  &45588953(45558663.1) \\
regi\_100  &92940074(91648450.48)  &93650192(\textbf{92642163.28})  &\textbf{93682151}(92599864.2) \\
	\hline
	\end{tabular}\label{tab:SAND-CATS-against-incomplete}
\end{table}

\begin{table} [h]\tiny
\setlength{\tabcolsep}{1.5pt}
\renewcommand{\arraystretch}{1.25}
	\caption{Results on SAND and CATS againt Complete Solvers}
    \centering
\begin{tabular}{|l|l|l|l|l|l|}
\hline
Graph &TSM-MWC &TRSC &Graph &TSM-MWC & TRSC \\
	& $w$& $w_{max} (w_{avg})$& & $w$ &  $w_{max} (w_{avg})$\\
\hline
\hline
Decay\_100 &\textbf{86372100}$^*$ &85510000(84706031.0)  & arbi\_40&40460122$^*$ &40460122(40460122.0)\\
Decay\_200 &\textbf{159676600} &151727300(147776429.0) &arbi\_100 &\textbf{89772104}$^*$ &89772104(87617454.9) \\
\cline{4-6}
Decay\_300&\textbf{222341300} &214579300(208712490.0) & matc\_80 &4544376 &\textbf{5081870}(\textbf{5070041.87})\\
\cline{4-6}
Decay\_400 &\textbf{267417100} &258063500(251908678.0) & path\_40 &\textbf{248707}$^*$ &248236(247778.48)\\
Decay\_500 &\textbf{326151300} &321463500(311769095.0) & path\_100 &\textbf{367045} &364972(361397.42) \\
\hline
uni\_200\_10 &11668500 &\textbf{12242300}(\textbf{12242300.0}) &regi\_40 &45588953$^*$ &45588953(45558663.1)\\
uni\_300\_10 &14745800 &\textbf{17201100}(\textbf{17201100.0}) & regi\_100 &\textbf{94017022}$^*$ &93682151(92599864.2) \\
\cline{4-6}
uni\_400\_10 &16872700 &\textbf{22022300}(\textbf{22022300.0}) &sche\_40 &546928 &\textbf{826022}(\textbf{826022.0}) \\
uni\_500\_10 &20499200 &\textbf{26564200}(\textbf{26553017.0}) &sche\_100 &1537860 &\textbf{1672303}(\textbf{1672303.0}) \\
	\hline
	\end{tabular}\label{tab:SAND-CATS-against-complete}
\end{table}

\begin{center}
\begin{table} [h]\scriptsize
\setlength{\tabcolsep}{11.0pt}
\renewcommand{\arraystretch}{1.0}
	\caption{Results on Kidney Exchange Schemes}
    \centering
\begin{tabular}{|l|l|l|}
\hline
Graph &TSM-MWC &TRSC \\
	& $w$& $w_{max} (w_{avg})$\\
\hline
\hline
081.wclq &1650240634894 &\textbf{1650240634895} (\textbf{1650240634894.14}) \\
091.wclq &1306441900045 &\textbf{1306441900046} (\textbf{1306441900045.87}) \\
092.wclq &\textbf{1581403750408} &1581403750408 (1569034576044.46) \\
095.wclq &1375228477453 &\textbf{1375245246480} (\textbf{1375245246479.88}) \\
096.wclq &1306374823942 &\textbf{1375094325251} (\textbf{1357914449924.0}) \\
097.wclq &1375144632329 &\textbf{1375144632330} (\textbf{1375144632329.29}) \\
099.wclq &1237722398734 &\textbf{1237722398735} (\textbf{1237722398735.0}) \\
100.wclq &1512701018123 &\textbf{1512701018124} (\textbf{1512701018123.9}) \\
	\hline
	\end{tabular}\label{tab:KES}
\end{table}
\end{center}

\begin{center}
\begin{table} [h]\tiny
\setlength{\tabcolsep}{1.8pt}
\renewcommand{\arraystretch}{1.0}
	\caption{Results on RAG-REF Graphs}
    \centering
\begin{tabular}{|l|l|l|l|l|l|}
\hline
Graph &TSM-MWC &TRSC & Graph &TSM-MWC &TRSC  \\
	& $w$& $w_{max} (w_{avg})$ & & $w$& $w_{max} (w_{avg})$\\
\hline
\hline
ref-60-230-0.clq &502 &\textbf{505}(\textbf{504.66})&ref-60-500-0.clq &651 &\textbf{696}(\textbf{696.0}) \\
ref-60-230-1.clq &501 &\textbf{506}(\textbf{505.0})&ref-60-500-1.clq &681 &\textbf{709}(\textbf{709.0}) \\
ref-60-230-2.clq &492 &\textbf{524}(\textbf{523.98})&ref-60-500-2.clq &650 &\textbf{701}(\textbf{694.07}) \\
ref-60-230-3.clq &492 &\textbf{502}(\textbf{502.0})&ref-60-500-3.clq &673 &\textbf{716}(\textbf{716.0}) \\
ref-60-230-4.clq &502 &\textbf{504}(\textbf{503.57})&ref-60-500-4.clq &627 &\textbf{690}(\textbf{689.99}) \\
ref-60-230-5.clq &500 &\textbf{503}(\textbf{502.81})&ref-60-500-5.clq &660 &\textbf{714}(\textbf{714.0})  \\
ref-60-230-6.clq &503 &\textbf{505}(\textbf{505.0})&ref-60-500-6.clq &669 &\textbf{715}(\textbf{715.0}) \\
ref-60-230-7.clq &503 &\textbf{506}(\textbf{504.94}) &ref-60-500-7.clq &657 &\textbf{692}(\textbf{692.0})\\
ref-60-230-8.clq &489 &\textbf{494}(\textbf{493.89}) &ref-60-500-8.clq &659 &\textbf{714}(\textbf{713.99})\\
ref-60-230-9.clq &481 &\textbf{526}(\textbf{525.66})&ref-60-500-9.clq &642 &\textbf{704}(\textbf{696.72}) \\
ref-60-300.clq &590 &\textbf{599}(\textbf{599.0})&ref-60-500.clq &679 &\textbf{704}(\textbf{704.0}) \\

	\hline
	\end{tabular}\label{tab:REF}
\end{table}
\end{center}

As to the other instances, our evaluation results are divided into two parts:
\begin{enumerate}
\item DIMACS and BHOSLIB graphs (97 instances); 
\item a list of \textbf{challenging} graphs including: (1) WDP graphs from the SAND and the CATS groups (30 instances), (2) KES graphs (50 instances) and (3) REF graphs from the RAG-REF group (29 instances).
\end{enumerate}

\subsection{DIMACS and BHOSLIB Graphs}
Experimental results show that TRSC significantly outperforms TSM-MWC, LSCC and RRWL in terms of average solution quality. 
For the sake of space, we exclude those graphs containing less than 1,400 vertices, but we keep one graph \texttt{MANN\_a45} which contains less than 1,400 vertices, because TSM-MWC outperforms TRSC on this instance. 
The detailed results are shown in Table \ref{tab:vertex-weighted-BHOSLIB-DIMACS}.

In order to show the gap between TSM-MWC and TRSC, we extended the cutoff to be 72 hours and tested TSM-MWC again. 
The solution quality of TSM-MWC on most instances in this table still falls behind that of TRSC significantly.

\subsection{Challenging Graphs}
Judging by recent SAT/MaxSAT Competitions\footnote{http://www.satcompetition.org/;http://www.maxsat.udl.cat/}, there is a prevailing hypothesis that exact solvers perform better on benchmarks from real-world applications.
Moreover, TSM-MWC has proved to be state-of-the-art over the graphs from practical applications \cite{JiangLLM18}.
So we mainly compare our solver with TSM-MWC here.
Considering that the SAND and CATS benchmarks have not been used often, we also tested LSCC and RRWL on them.

\subsubsection{SAND and CATS}
Table \ref{tab:SAND-CATS-against-incomplete} shows the comparisons between TRSC and state-of-the-art incomplete solvers.
Table \ref{tab:SAND-CATS-against-complete} shows the comparisons between TRSC and TSM-MWC.
From these tables, we can see that TRSC significantly outperforms LSCC and RRWL.
Also TRSC is complementary with TSM-MWC. Both solvers perform as well as each other.

\subsubsection{KES and REF}
Tables \ref{tab:KES} shows the comparisons between TRSC and TSM-MWC on Kidney-exchange Schemes, and Tables \ref{tab:REF} shows respective results on Research Excellence Framework.
From these tables, we can see that TRSC significantly outperforms TSM-MWC.
Moreover, these results refute the prevailing hypothesis that local search algorithms are less well suited for application instances.

\subsection{Restart Periods}

We selected 8 instances from different benchmarks, and evaluated the restart periods of LSCC, RRWL and TRSC. We used 3600s as the cutoff and seeds from 1 to 10. 
The results are in Table \ref{tab:restart-period}. 
For instance, on \texttt{100.wclq} in the Kidney-Exchange Schemes benchmark, RRWL restarts every 29,033 steps while TRSC restarts every 34,685 steps on average.
Notice that LSCC always restarts every 4,000 steps simply because of its default parameter setting \cite{DBLP:conf/aaai/WangCY16}.
In Table \ref{tab:restart-period}, we can find that:
\begin{enumerate}
\item the restart periods of RRWL and TRSC vary significantly from instance to instance;
\item TRSC usually has a longer restart period than RRWL, $\ie$, TRSC usually restarts less frequently than RRWL.
\end{enumerate}
This is consistent with our expectations since TRSC employ stronger tabu and more conservative restart strategies.

\begin{table} [h]\small
\setlength{\tabcolsep}{3.3pt}
\renewcommand{\arraystretch}{1.25}
	\caption{Restart Periods of LSCC, RRWL and TRSC}
    \centering
\begin{tabular}{|l|l|l|l|}
\hline
Graph  &LSCC &RRWL &TRSC \\
\hline
\hline
100.wclq  &4,000      &29,033    &34,685 \\
arbitrary\_40.txt  &4,000      &52          &234 \\
Decay2000\_500.txt &4,000       &952        &1,424 \\
frb-80-33-5.clq  &4,000     &1,289     &2,313 \\
MANN\_a81.clq  &4,000     &40,055   &157,602 \\
ref-60-500.clq &4,000     &3,060     &2,988 \\
scheduling\_100.txt  &4,000    &1,389      &8,663 \\
uniform2000\_500\_10.txt  &4,000    &425         &608  \\
	\hline
	\end{tabular}\label{tab:restart-period}
\end{table}

\section{Conclusions and Future Work}

In this paper, we advanced both tabu and restart strategies based on the notion of a local search scenario, and developed a local search search MWC solver called TRSC. 
TRSC outperforms several state-of-the-art solvers by extensive experiments including those on the two influential benchmarks of BHOSLIB and DIMACS.
Moreover, the reported results refute the prevailing hypothesis that local search algorithms are less well suited for application graphs.

As for future work, we will study variants of the tabu and restart strategies in other combinatorial optimization problems like maximum satisfiability and minimum vertex cover, as these two strategies are fundamental ones for local search.
Currently we are investigating whether these strategy are also effective in the classic maximum clique problem.

\bibliographystyle{named}
\bibliography{ijcai18}

\begin{thebibliography}{}

\bibitem[\protect\citeauthoryear{Battiti and
  Protasi}{2001}]{battiti2001reactive}
Roberto Battiti and Marco Protasi.
\newblock Reactive local search for the maximum clique problem.
\newblock {\em Algorithmica}, 29(4):610--637, 2001.

\bibitem[\protect\citeauthoryear{Brendel and
  Todorovic}{2010}]{DBLP:conf/nips/BrendelT10}
William Brendel and Sinisa Todorovic.
\newblock Segmentation as maximum-weight independent set.
\newblock In {\em Advances in Neural Information Processing Systems 23: 24th
  Annual Conference on Neural Information Processing Systems 2010. Proceedings
  of a meeting held 6-9 December 2010, Vancouver, British Columbia, Canada.},
  pages 307--315, 2010.

\bibitem[\protect\citeauthoryear{Brendel \bgroup \em et al.\egroup
  }{2011}]{DBLP:conf/cvpr/BrendelAT11}
William Brendel, Mohamed~R. Amer, and Sinisa Todorovic.
\newblock Multiobject tracking as maximum weight independent set.
\newblock In {\em The 24th {IEEE} Conference on Computer Vision and Pattern
  Recognition, {CVPR} 2011, Colorado Springs, CO, USA, 20-25 June 2011}, pages
  1273--1280, 2011.

\bibitem[\protect\citeauthoryear{Cai and Lin}{2016}]{DBLP:conf/ijcai/CaiL16}
Shaowei Cai and Jinkun Lin.
\newblock Fast solving maximum weight clique problem in massive graphs.
\newblock In {\em Proceedings of the Twenty-Fifth International Joint
  Conference on Artificial Intelligence, {IJCAI} 2016, New York, NY, USA, 9-15
  July 2016}, pages 568--574, 2016.

\bibitem[\protect\citeauthoryear{Cai \bgroup \em et al.\egroup
  }{2011}]{DBLP:journals/ai/CaiSS11}
Shaowei Cai, Kaile Su, and Abdul Sattar.
\newblock Local search with edge weighting and configuration checking
  heuristics for minimum vertex cover.
\newblock {\em Artif. Intell.}, 175(9-10):1672--1696, 2011.

\bibitem[\protect\citeauthoryear{Fan \bgroup \em et al.\egroup
  }{2016}]{DBLP:conf/ausai/FanLMWSS16}
Yi~Fan, Chengqian Li, Zongjie Ma, Lian Wen, Abdul Sattar, and Kaile Su.
\newblock Local search for maximum vertex weight clique on large sparse graphs
  with efficient data structures.
\newblock In {\em {AI} 2016: Advances in Artificial Intelligence - 29th
  Australasian Joint Conference, Hobart, TAS, Australia, December 5-8, 2016,
  Proceedings}, pages 255--267, 2016.

\bibitem[\protect\citeauthoryear{Fan \bgroup \em et al.\egroup
  }{2017a}]{DBLP:conf/ijcai/FanLLMLS17}
Yi~Fan, Nan Li, Chengqian Li, Zongjie Ma, Longin~Jan Latecki, and Kaile Su.
\newblock Restart and random walk in local search for maximum vertex weight
  cliques with evaluations in clustering aggregation.
\newblock In {\em Proceedings of the Twenty-Sixth International Joint
  Conference on Artificial Intelligence, {IJCAI} 2017, Melbourne, Australia,
  August 19-25, 2017}, pages 622--630, 2017.

\bibitem[\protect\citeauthoryear{Fan \bgroup \em et al.\egroup
  }{2017b}]{DBLP:conf/ictai/FanMKCRLL17}
Yi~Fan, Zongjie Ma, Kaile Su, Chengqian Li, Cong Rao, Ren-Hau Liu, and
  Longin~Jan Latecki.
\newblock Efficient local search for maximum weight cliques in large graphs.
\newblock In {\em 2017 {IEEE} 29th International Conference on Tools with
  Artificial Intelligence, Boston, MA, USA, November 6-8, 2017}, page to
  appear, 2017.

\bibitem[\protect\citeauthoryear{Fang \bgroup \em et al.\egroup
  }{2016}]{DBLP:journals/jair/FangLX16}
Zhiwen Fang, Chu{-}Min Li, and Ke~Xu.
\newblock An exact algorithm based on maxsat reasoning for the maximum weight
  clique problem.
\newblock {\em J. Artif. Intell. Res.}, 55:799--833, 2016.

\bibitem[\protect\citeauthoryear{Hoos and
  St{\"{u}}tzle}{2007}]{DBLP:reference/crc/HoosS07a}
Holger~H. Hoos and Thomas St{\"{u}}tzle.
\newblock Stochastic local search.
\newblock In {\em Handbook of Approximation Algorithms and Metaheuristics.}
  2007.

\bibitem[\protect\citeauthoryear{Jiang \bgroup \em et al.\egroup
  }{2017}]{DBLP:conf/aaai/JiangLM17}
Hua Jiang, Chu{-}Min Li, and Felip Many{\`{a}}.
\newblock An exact algorithm for the maximum weight clique problem in large
  graphs.
\newblock In {\em Proceedings of the Thirty-First {AAAI} Conference on
  Artificial Intelligence, February 4-9, 2017, San Francisco, California,
  {USA.}}, pages 830--838, 2017.

\bibitem[\protect\citeauthoryear{Jiang \bgroup \em et al.\egroup
  }{2018}]{JiangLLM18}
Hua Jiang, Chu{-}Min Li, and Felip~Many{\`{a}} Yanli~Liu.
\newblock A two-stage maxsat reasoning approach for the maximum weight clique
  problem.
\newblock In {\em Proceedings of the Thirty-Second {AAAI} Conference on
  Artificial Intelligence, February 2-7, 2018, New Orleans, Louisiana, {USA.}},
  page To appear, 2018.

\bibitem[\protect\citeauthoryear{Johnson and
  Trick}{1996}]{Johnson:1996:CCS:548182}
David~J. Johnson and Michael~A. Trick, editors.
\newblock {\em Cliques, Coloring, and Satisfiability: Second DIMACS
  Implementation Challenge, Workshop, October 11-13, 1993}.
\newblock American Mathematical Society, Boston, MA, USA, 1996.

\bibitem[\protect\citeauthoryear{Lau and Goh}{2002}]{DBLP:conf/ictai/LauG02}
Hoong~Chuin Lau and Yam~Guan Goh.
\newblock An intelligent brokering system to support multi-agent web-based
  4th-party logistics.
\newblock In {\em 14th {IEEE} International Conference on Tools with Artificial
  Intelligence {(ICTAI} 2002), 4-6 November 2002, Washington, DC, {USA}}, page
  154, 2002.

\bibitem[\protect\citeauthoryear{Leyton{-}Brown \bgroup \em et al.\egroup
  }{2000}]{DBLP:conf/sigecom/Leyton-BrownPS00}
Kevin Leyton{-}Brown, Mark Pearson, and Yoav Shoham.
\newblock Towards a universal test suite for combinatorial auction algorithms.
\newblock In {\em {EC}}, pages 66--76, 2000.

\bibitem[\protect\citeauthoryear{Li and Latecki}{2012}]{DBLP:conf/nips/LiL12}
Nan Li and Longin~Jan Latecki.
\newblock Clustering aggregation as maximum-weight independent set.
\newblock In {\em Advances in Neural Information Processing Systems 25: 26th
  Annual Conference on Neural Information Processing Systems 2012. Proceedings
  of a meeting held December 3-6, 2012, Lake Tahoe, Nevada, United States.},
  pages 791--799, 2012.

\bibitem[\protect\citeauthoryear{McCreesh \bgroup \em et al.\egroup
  }{2017}]{DBLP:conf/cp/McCreeshPST17}
Ciaran McCreesh, Patrick Prosser, Kyle Simpson, and James Trimble.
\newblock On maximum weight clique algorithms, and how they are evaluated.
\newblock In {\em Principles and Practice of Constraint Programming - 23rd
  International Conference, {CP} 2017, Melbourne, VIC, Australia, August 28 -
  September 1, 2017, Proceedings}, pages 206--225, 2017.

\bibitem[\protect\citeauthoryear{Nogueira \bgroup \em et al.\egroup
  }{2017}]{Nogueira2017}
Bruno Nogueira, Rian G.~S. Pinheiro, and Anand Subramanian.
\newblock A hybrid iterated local search heuristic for the maximum weight
  independent set problem.
\newblock {\em Optimization Letters}, Mar 2017.

\bibitem[\protect\citeauthoryear{\"{O}sterg{\aa}rd}{2001}]{Ostergard:2001:NAM:%
766502.766504}
Patric R.~J. \"{O}sterg{\aa}rd.
\newblock A new algorithm for the maximum-weight clique problem.
\newblock {\em Nordic J. of Computing}, 8(4):424--436, December 2001.

\bibitem[\protect\citeauthoryear{Pullan}{2008}]{DBLP:journals/heuristics/Pulla%
n08}
Wayne~J. Pullan.
\newblock Approximating the maximum vertex/edge weighted clique using local
  search.
\newblock {\em J. Heuristics}, 14(2):117--134, 2008.

\bibitem[\protect\citeauthoryear{Sandholm}{2002}]{DBLP:journals/ai/Sandholm02}
Tuomas Sandholm.
\newblock Algorithm for optimal winner determination in combinatorial auctions.
\newblock {\em Artif. Intell.}, 135(1-2):1--54, 2002.

\bibitem[\protect\citeauthoryear{Shimizu \bgroup \em et al.\egroup
  }{2012}]{shimizu2012some}
Satoshi Shimizu, Kazuaki Yamaguchi, Toshiki Saitoh, and Sumio Masuda.
\newblock Some improvements on kumlander's maximum weight clique extraction
  algorithm.
\newblock In {\em Proceedings of World Academy of Science, Engineering and
  Technology}, number~72, page 948. World Academy of Science, Engineering and
  Technology (WASET), 2012.

\bibitem[\protect\citeauthoryear{Wang \bgroup \em et al.\egroup
  }{2016}]{DBLP:conf/aaai/WangCY16}
Yiyuan Wang, Shaowei Cai, and Minghao Yin.
\newblock Two efficient local search algorithms for maximum weight clique
  problem.
\newblock In {\em Proceedings of the Thirtieth {AAAI} Conference on Artificial
  Intelligence, February 12-17, 2016, Phoenix, Arizona, {USA.}}, pages
  805--811, 2016.

\bibitem[\protect\citeauthoryear{Wu \bgroup \em et al.\egroup
  }{2012}]{DBLP:journals/anor/WuHG12}
Qinghua Wu, Jin{-}Kao Hao, and Fred Glover.
\newblock Multi-neighborhood tabu search for the maximum weight clique problem.
\newblock {\em Annals {OR}}, 196(1):611--634, 2012.

\bibitem[\protect\citeauthoryear{Xu \bgroup \em et al.\egroup
  }{2005}]{DBLP:conf/ijcai/XuBHL05}
Ke~Xu, Fr{\'{e}}d{\'{e}}ric Boussemart, Fred Hemery, and Christophe Lecoutre.
\newblock A simple model to generate hard satisfiable instances.
\newblock In {\em IJCAI-05, Proceedings of the Nineteenth International Joint
  Conference on Artificial Intelligence, Edinburgh, Scotland, UK, July 30 -
  August 5, 2005}, pages 337--342, 2005.

\bibitem[\protect\citeauthoryear{Yamaguchi and Masuda}{2008}]{yamaguchi2008new}
Kazuaki Yamaguchi and Sumio Masuda.
\newblock A new exact algorithm for the maximum weight clique problem.
\newblock {\em ITC-CSCC: 2008}, pages 317--320, 2008.

\bibitem[\protect\citeauthoryear{Zhou \bgroup \em et al.\egroup
  }{2017}]{DBLP:journals/eor/ZhouHG17}
Yi~Zhou, Jin{-}Kao Hao, and Adrien Go{\"{e}}ffon.
\newblock {PUSH:} {A} generalized operator for the maximum vertex weight clique
  problem.
\newblock {\em European Journal of Operational Research}, 257(1):41--54, 2017.

\end{thebibliography}

\end{document}